\documentclass{article}

\PassOptionsToPackage{numbers, compress}{natbib}

\usepackage[preprint]{neurips_2024}  

\usepackage{float}
\usepackage{arydshln}
\usepackage{wrapfig}
\usepackage{xspace}
\usepackage[utf8]{inputenc} 
\usepackage[T1]{fontenc}    
\usepackage{url}            
\usepackage{booktabs}       
\usepackage{amsfonts}       
\usepackage{microtype}
\usepackage{nicefrac}       
\usepackage{tabularx} 
\usepackage{microtype}      
\usepackage{xcolor}         
\usepackage{colortbl}
\usepackage{subcaption}
\usepackage[inline]{enumitem}
\usepackage{graphicx}
\usepackage{multirow}

\usepackage{amsmath}
\usepackage{makecell} 
\usepackage{amsfonts,amssymb}
\usepackage{pifont}       
\usepackage{bbding}       
\usepackage{fontawesome}  
\usepackage{tcolorbox}
\usepackage{adjustbox}
\usepackage{hyperref}
\usepackage{listings}

\newcommand{\eg}{\emph{e.g.,}\xspace}

\newcommand{\ignore}[1]{}

\newcommand{\model}{ASFM\xspace}
\newcommand{\fullmodel}{\textbf{A}gent-based \textbf{S}imulated \textbf{F}inancial \textbf{M}arket\xspace}

\newcommand{\instrctionsize}{\normalsize}

\title{Simulating Financial Market via \\ Large Language Model based Agents}

\author{\textbf{Shen Gao}\textsuperscript{1}, \textbf{Yuntao Wen}\textsuperscript{1}, \textbf{Minghang Zhu}\textsuperscript{2}, \textbf{Jianing Wei}\textsuperscript{1},\\ \textbf{Yuhan Cheng}\textsuperscript{2},
\textbf{Qunzi Zhang}\textsuperscript{2}, \textbf{Shuo Shang}\textsuperscript{1}\thanks{\, Corresponding author.}\\
  \textsuperscript{1} University of Electronic Science and Technology of China\\
  \textsuperscript{2} Shandong University\\
  \texttt{shengao@uestc.edu.cn yuntaowenx@gmail.com mhzhu@mail.sdu.edu.cn} \\
  \texttt{jedi.shang@gmail.com}}

\begin{document}

\maketitle

\begin{abstract}
  Most economic theories typically assume that financial market participants are fully rational individuals and use mathematical models to simulate human behavior in financial markets. 
  However, human behavior is often not entirely rational and is challenging to predict accurately with mathematical models. 
  In this paper, we propose \fullmodel (\model), which first constructs a simulated stock market with a real order matching system.
  Then, we propose a large language model based agent as the stock trader, which contains the profile, observation, and tool-learning based action module.
  The trading agent can comprehensively understand current market dynamics and financial policy information, and make decisions that align with their trading strategy. 
  In the experiments, we first verify that the reactions of our \model are consistent with the real stock market in two controllable scenarios.
  In addition, we also conduct experiments in two popular economics research directions, and we find that conclusions drawn in our \model align with the preliminary findings in economics research.
  Based on these observations, we believe our proposed \model provides a new paradigm for economic research.
  
\end{abstract}

\section{Introduction}


For a long time, market institutions such as fund companies and policy-making institutions such as central banks have been seeking a type of theoretical model capable of modeling economic operations to effectively analyze the outcomes of policy implementation and simulate the future functioning of the economy. 
Economists have long struggled to solve the problem of modeling complex human behavior. 
However, a fundamental assumption in traditional economic theory is that participants in economic markets are rational. 
For the past 100 years, economic research has followed certain economic assumptions, mostly based on rationality, with a few considering some features of irrational behavior. 
Nonetheless, these methods cannot fully capture complex human behavior.

Since it is hard to fully model the operating principles of the financial market, economic regulatory policies cannot achieve the expected effects by using the prediction based on economic theory. 
Therefore, some economists have proposed behavioral economics~\cite{rick2008role, lavecchia2016behavioral}, which aims to explore the social, cognitive, and emotional factors influencing economic decisions made by individuals and groups. 
In recent years, this field has become a hot topic in economics, exemplified by the awarding of the 2017 Nobel Prize in Economics to Richard Thaler for his significant contributions to behavioral economics and behavioral finance~\cite{barberis2003survey,kahneman1986fairness}. 
Since humans are not fully rational and their behavior under irrational conditions is difficult to predict, it still remains challenging to model complex human behavior using mathematical models.

Additionally, some economists have proposed experimental economics~\cite{smith1976experimental,davis2021experimental}, which involves using human experiments to test different economic theories and new market mechanisms. 
Experimental economics helps researchers understand the principles of market and trading system operations.
However, the most commonly used method~\cite{smith1962experimental,ariely2007psychology,kahneman2002foundations} in experimental economics involves employing human subjects to simulate real trading in a laboratory environment.
This approach is not only costly but also difficult to scale for large-scale and long-term experiments. 

The large language models (LLMs) based agents have emerged as a novel research direction in recent years~\cite{park2023generative,liu2023training,sumers2023cognitive}. 
LLM-based agents typically encompass the following modules: agent profile definition~\cite{park2023generative}, agent memory construction~\cite{lee2023stockemotions,wang2024survey}, agent planning capabilities~\cite{xie2024travelplanner}, and executable actions~\cite{zhang2023building}. 
Due to the reasoning and external tool usage capabilities, researchers begin exploring how multiple agents could collaborate to accomplish more complex tasks~\cite{qin2023toolllm,hsieh2023tool}.
These works have shown that large language models exhibit a high degree of consistency in simulating human behavior. 

In this paper, we propose \fullmodel (\model), a stock market simulation framework based on language model agents. 
First, we constructed a simulated stock trading market, encompassing most industry sectors present in the real financial market and implementing an order-matching trading mechanism identical to that of real markets. 
Subsequently, we developed several stock trading agents with diverse profiles and trading strategies. 
We equip each agent with a market observation module and stock trading capabilities. 
Additionally, we also simulate the proportions of various participants in the real financial market, making the trading data in \model more closely resemble that of a real financial market.

In the experiment, we first validated two controlled scenarios within the simulated market: the effects of interest rate changes and the impact of inflation shock on the stock market. 
Through these two experiments, we confirm that our proposed \model can effectively simulate the principles of a real market, thereby verifying the accuracy of the proposed simulation methods. 
Based on \model, we further explored two crucial issues in economic research: the impact of trader behavior bias and the large trader impact. 
By conducting long-term simulation experiments in these two scenarios, we can find that \model shows consistency with the preliminary findings in recent economic research.

\noindent Our contributions are as follows:


\noindent $\bullet$ Our paper reveals that by integrating LLMs with the mechanisms of economic and financial operations, our proposed \model and technical route of policy analysis are expected to undergo a comprehensive revolution for economic research.

\noindent $\bullet$ We construct an actual trading matching system and employ several simulated listed companies consistent with the real market.

\noindent $\bullet$ We propose an LLM-based stock trading agent, that can observe market fluctuations and receive external economic news, using tool learning to perform stock trading in the simulated market based on their observations.

\noindent $\bullet$ We demonstrate the consistent observation between our simulated financial market and the real-world market by conducting quantitative simulation experiments in several complex scenarios.

\section{\model Framework}

\begin{figure}[h]
    \centering
      \includegraphics[width=\columnwidth]{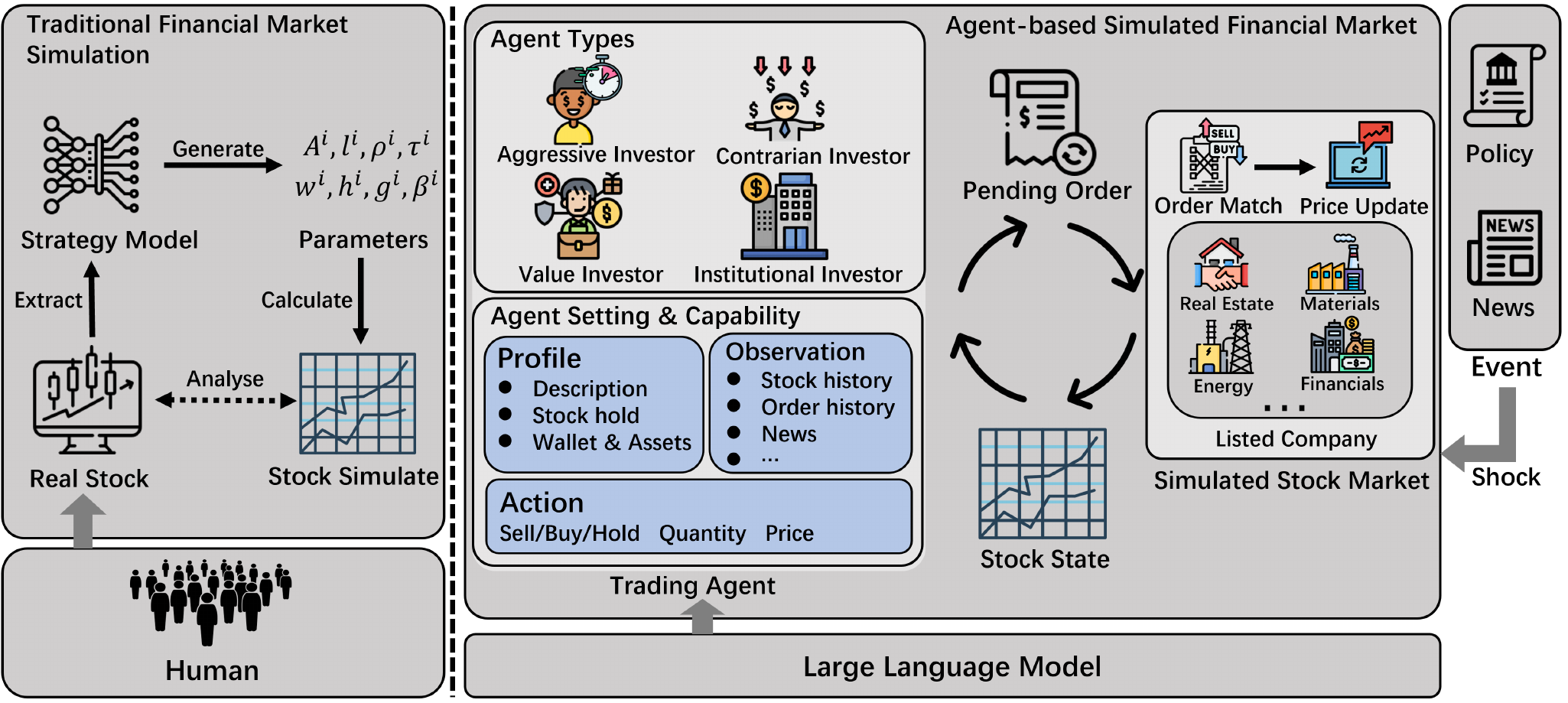}
      \caption{Comparison between our proposed \fullmodel (\model) (shown in the right) and traditional economic research paradigm (shown in the left). \model does not require any human annotator involved and only uses the LLM-based agents to conduct micro-financial market simulation.}
      \label{fig:model}
\end{figure}

\subsection{Problem Formulation}

First, we define the simulated stock market as $\mathcal{M}=\{\mathcal{M}_1, \dots, \mathcal{M}_m\}$, which consists of $m$ listed companies. 
Each listed company $\mathcal{M}_i = \{\mathcal{S}_i, \mathcal{H}_i^T\}$ is composed of its business description $\mathcal{S}_i$ and the stock prices over the past $T$ days $\mathcal{H}_i^T$.
To simulate the stock market, we construct $n$ trader agents $\mathcal{A} = \{\mathcal{A}_1, \dots, \mathcal{A}_n\}$. 
Each agent $\mathcal{A}_i = \{\mathcal{D}_i, \mathcal{U}_i, \mathcal{W}_i\}$ consists of its trading strategy $\mathcal{D}_i$, the stocks it holds $\mathcal{U}_i$ with quantities, and the cash it owns $\mathcal{W}_i$.
After observing stock price changes and current economic news $P$, the agents complete stock transactions by selecting tools from the toolset $\mathcal{T}$ and generate the tool execution code $C$. 
The toolset $\mathcal{T}$ includes three tools: Sell, Buy, and Hold. 
Each tool accepts three parameters: stock code, quantity, and price.
Our task is to enable the LLM-based agents to choose the appropriate tool in $\mathcal{T}$ based on their profile $\mathcal{A}_i$ and observe market conditions, finally generating the code $C$ for trading decisions and placing orders in the trading system.

\subsection{Overall Framework}

In this section, we detail the \fullmodel (\model).
An overview of \model is shown in Figure~\ref{fig:model}.
\model has two main parts:

(1) \textbf{Simulated Stock Market.} To construct a simulated stock market, we first define several publicly listed companies and describe their main business activities. 
Then, we constructed an order-matching system according to the rules of the real-world stock market. 
Finally, we implemented a stock price updating algorithm to update the stock prices daily based on trading activities.

(2) \textbf{LLM-based Trading Agent.} We first define each agent's profile and initial capital. 
For an agent to complete stock trades, it should observe the stock market trends before trading in the market. 
Therefore, we implement an observation module to enable the agent to understand stock price fluctuations. 
Finally, we enabled the agent to interact with the simulated stock market by using a tool-learning paradigm.

\subsection{Simulated Stock Market}

\subsubsection{Listed Company}

In the real-world stock market, numerous publicly listed companies issue and trade stocks in the stock market.
Therefore, we first define what companies are listed in our simulated stock market. 
To better simulate real-world market, we created several simulated companies according to the industry sector distribution proportions observed in the real-world stock market. 
When defining each stock, we initially provide its historical price data $\mathcal{H}_i^T$ and a description of the company's main business activities $\mathcal{S}_i$. 
During the initialization of the simulated stock market, we used real stock data as the starting prices for the first five days. 
We show an example listed company description below:

\begin{tcolorbox}[colback=black!1!white,colframe=black!57!white,boxsep=1pt,left=1pt,right=1pt,top=1pt,bottom=1pt]
    \instrctionsize
    \textbf{Company Name}: IT008\\
    \textbf{Sector}: Information Technology\\
    \textbf{Main Business $\mathcal{S}_i$}: A leading software development and information technology services provider\\
    \textbf{Historical Prices $\mathcal{H}_i^T$}: [80.00, 79.75, 80.50, 81.25, 81.75...]\\
    \textbf{Initial Prices}: Based on the first five days' real stock data
\end{tcolorbox}

\subsubsection{Order Matching System}
An order matching system is a system that matches buy and sell orders for a stock market.
In this paper, we use two types of transaction matching mechanisms: opening order matching and continuous order matching.

Opening matching orders refer to the process where sellers place sell orders and buyers place buy orders in the trading market. 
The market determines the transaction price based on the principles of price priority and time priority~\cite{OHara1995MarketMT}. 
For each stock, we prioritize matching the lowest sell orders and the highest buy orders. 
The transaction price is the average price of the sell and buy prices. 
When all the stock quantity of the lowest sell order or all the stock quantity of the highest buy order has been traded, then proceed to match the second lowest sell order or the second highest order, and repeat this step until trading stops.
There are two conditions for the trading stop: (1) the buy orders or sell orders on one side are completely exhausted, and (2) the price of the lowest sell order is higher than the price of the highest buy order.


Continuous order matching refers to the process where, after the initial orders at the opening have not been fully executed, the unmatched orders are made available as indicative orders that agents can observe. 
Agents can adjust their bids based on these indicative orders and place new orders.

\paragraph{Stock Price Updating Strategy}
At the end of each trading day, we use the average transaction price of each stock as the closing price for that day. 
We then use the closing price of each stock to calculate and update each agent's total assets based on the value of the stocks they hold.

\subsection{LLM-based Trading Agent}

\subsubsection{Agent Profile}\label{sec:agent-profile}

In real stock trading, different individuals adopt various investment strategies, leading to significantly different trading behaviors. 
To better simulate a real stock market, we first define the investment strategy $\mathcal{D}_i$ of the $i$-th agent when constructing the trading agents and then inform the agent of the detailed trading preference of this strategy. 
In this paper, we define four types of different investment strategies: value investors, institutional investors, contrarian investors, and aggressive investors. 
For example, when defining a value investing agent, we specify in its profile that it should focus on finding stocks whose market prices are below their intrinsic values and identify and invest in high-quality companies undervalued by the market. 
More detailed profiles are provided in Appendix~\ref{sec:appendix.A1}.

Additionally, we allocated initial capital to each agent.
To help agents understand their current holdings, we designed an independent account management mechanism for each agent. 
After each transaction, our system calculates the agent's remaining cash balance $\mathcal{W}_i$, quantities of held stocks $\mathcal{U}_i$, and rate of return, among other common statistical data. 
Based on this information, we construct the trader agent prompt $\mathcal{I}_a$ as follows:
\begin{tcolorbox}
[colback=black!1!white,colframe=black!57!white,boxsep=1pt,left=1pt,right=1pt,top=1pt,bottom=1pt]
    \instrctionsize
    You are an investor in the stock market, and $\mathcal{D}_i$ \{strategy\_description\}\\ 
    \text{\textbf{[Wallet $\mathcal{W}_i$]}}
    \{wallet\_cash\}\\
    \text{\textbf{[Stock $\mathcal{U}_i$]}}
    \{stocks\_hold\}
\end{tcolorbox}

\subsubsection{Agent Observation}
To help agents better understand the current market fluctuations, we prompt the agent to observe stock trading data and current economic policies. 
We use the observation prompt $\mathcal{I}_o$ to the agent with three types of information: (1) the stock price over the past 15 days, (2) the historical prices of the stock's order book, and (3) current economic policy news:
\begin{tcolorbox}
[colback=black!1!white,colframe=black!57!white,boxsep=1pt,left=1pt,right=1pt,top=1pt,bottom=1pt]
...Omit the description for the input information...\\
\text{\textbf{[Stock Market Situation]}} 
\{stocks\_company\}\{stocks\_history\_price\}\\
\text{\textbf{[Orders]}} 
\{orders\_history\}\\
\text{\textbf{[Economic News]}}
\{news\} \\
Please follow the steps below to generate the answer step by step:\\
...Omit the instructions for generating observation results...
\end{tcolorbox}
With these three types of information, agents can comprehend the stock's price movements and adjust their investment action based on the latest economic policy information.

\subsubsection{Agent Action}

Inspired by recent work on LLM tool learning~\cite{qin2023tool,gao2024confucius,zhuang2024toolqa}, we defined stock trading as three tools for the agent: buying stocks, selling stocks, and holding stocks. 
Specifically, the parameters for these tools are ``stock code, quantity, and price''. 
Finally, we use the profile prompt $\mathcal{I}_a$ and observation prompt $\mathcal{I}_o$ as the input of agent to produce the trading action execution:
\begin{equation}
    \mathcal{C} = \mathcal{A}(\mathcal{I}_a, \mathcal{I}_o, \mathcal{I}_t),
\end{equation}
where $\mathcal{A}$ is the LLM-based agent, $\mathcal{C}$ indicates the tool execution code, and $\mathcal{I}_t$ is the tool document and demonstration instruction.
Additionally, we limit each agent to perform no more than two buy or sell operation per stock each day.
\section{Experimental Setup}

\subsection{Evaluation Metrics}\label{sec:metrics}

To quantitatively evaluate the trading dynamics in our simulated stock market, we employed four commonly used stock market metrics:

(1) \textbf{Order Number (ON)}: This refers to the total number of buy and sell orders, reflecting the willingness of buyers and sellers at that price.

(2) \textbf{Order Execution Rate (OER)}: This is the ratio of order volume to the order quantity, reflecting the efficiency of order execution or the proportion of orders that result in trades.

(3) \textbf{Turnover Rate (TR)}: This is the ratio of stock trading volume to the number of shares outstanding over a certain period, measuring the liquidity and market activity of the stock.

(4) \textbf{Volatility (VO)}: This represents the standard deviation of the stock's rate of change, indicating the extent of price fluctuations and measuring the stock's price stability and risk level.

Due to the limited space, we omit the detailed definition for each metric.

\subsection{Ablation Models}

To verify the effectiveness of each module in \model, we also employ several ablation models:

(1) \textbf{\texttt{\model w/o Prof.}}: We use the same simple profile for each agent, which only instructs the agent to use the trade tools according to the observation results.

(2) \textbf{\texttt{\model w/o Obser.}}: We remove the observation module and only use the stock price of recent days as input to the agent.


\subsection{Simulated Stocks}

In our experiments, we referenced the distribution of listed companies in China's A-shares market and used the main business descriptions of real listed companies as a blueprint to create simulated companies covering 11 industry sectors. 
These sectors are energy, materials, industrial, consumer discretionary, consumer staples, healthcare, financial, information technology, telecommunication services, utilities, and real estate.

\subsection{Simulated Agents}\label{sec:agent-cate}

As shown in \S~\ref{sec:agent-profile}, we define 4 types of investors with different investment strategies.
In our experiments, we reference investors' distribution proportions in China's A-shares market. 
According to statistics of China's A-shares market, the proportions of value investors, institutional investors, contrarian investors, and aggressive investors in A-shares are approximately 2:1:1:1, respectively. Based on the composition of capital in China's A-share market, we also set the initial capital of each type of agent to 20K, 15K, 0.4K, and 6K respectively. 

\begin{table}[ht]
    \centering
    \caption{Evaluation metrics of different models.}
    \label{tab:ablation}
    \begin{tabular}{lcccc}
        \toprule
        & \textbf{ON} & \textbf{OER} & \textbf{TR} & \textbf{VO} \\
        \midrule
        \texttt{\model} & 1855 & 17.25\% & 62.26\% & 1.45\% \\
        \texttt{w/o Prof.} & 320$_{\downarrow{82.7\%}}$ & 13.98\%$_{\downarrow{19.0\%}}$ & 10.12\%$_{\downarrow{83.7\%}}$ & 0.37\% $_{\downarrow{74.5\%}}$\\
        \texttt{w/o Obser.} & 802$_{\downarrow{56.8\%}}$ & 28.30\%$_{\uparrow{64.1\%}}$ & 44.16\%$_{\downarrow{29.1\%}}$ & 3.24\%$_{\uparrow{123\%}}$ \\
        \midrule
        All Value Investors & 1021$_{\downarrow{45.0\%}}$ & 13.70\%$_{\downarrow{20.3\%}}$ & 23.91\%$_{\downarrow{61.6\%}}$ & 0.97\%$_{\downarrow{33.1\%}}$ \\
        All Aggressive Investors & 3837$_{\uparrow{107\%}}$ & 9.42\%$_{\downarrow{45.4\%}}$ & 168.87\%$_{\uparrow{171\%}}$ & 2.68\%$_{\downarrow{84.8\%}}$ \\
        \bottomrule
    \end{tabular}
\end{table}

\subsection{Simulated Scenarios}\label{sec:scenarios}

The ultimate goal of our framework is to simulate the real stock market's response to economic regulatory policies. 
Therefore, we first verify whether our \model can exhibit reactions consistent with real stock traders in some simple and controllable scenarios. 
We simulated the following two scenarios for the agents and observed their performance:

(1) \textbf{Federal cuts interest rates}: A Federal interest rate cut refers to the decision by the Federal Reserve's Board of Governors to adjust monetary policy by lowering the federal funds rate. 
A rate cut is a form of expansionary monetary policy, typically designed to reduce bank interest rates, thereby increasing the money supply and leading to a depreciation of the dollar. 
This type of policy generally has a positive impact on the stock market.

(2) \textbf{Inflation shock}: Central banks (\eg the Federal Reserve) usually set an inflation target (\eg 2\%) to maintain price stability and economic health. 
When the actual inflation rate deviates from this target, it triggers a series of shocks and chain reactions. 
In the experiment, we inform the agents of the current inflation rate by releasing news to agents.

The above two scenarios usually have a determinate influence on the stock market.
However, the real financial market is influenced by numerous factors, making accurate trending prediction challenging. 
Some factors are complex and remain widely studied issues in economics~\cite{alam2009relationship,clarida2000monetary, gokal2004relationship, boudoukh1993stock}. 
Given that the goal of our framework is to simulate the stock market under the influence of complex factors, it aims to assist policymakers in better understanding the potential impacts of their policies on the market.
We also select two currently popular research directions in economics, using our proposed \model framework to simulate these factors and investigate whether the conclusions drawn in our simulated environment align with the preliminary findings in economics research:

(1) \textbf{Trader with behavior bias}: In \S~\ref{sec:agent-cate}, we defined several agents with different trading strategies. 
To better simulate the real stock market, we used the actual proportions of investor types when constructing the agents. 
To study how stock market volatility varies with different proportions of investor types, we alter the investor type ratios and compare it with the financial market with real investor type ratios.

(2) \textbf{Large trader impact}: In macroeconomics, numerous studies~\cite{Corsetti2004Does, Kraft2011Large, BANNIER20051517} have shown that large traders can exercise a disproportionate influence on the likelihood and severity of a financial crisis by fomenting and orchestrating attacks against weakened currency pegs. 
This highlights the significant difference in market impact between large and small traders. 
Therefore, we explore how the amount of capital affects the final returns of the agent.

\section{Experimental Results}



\subsection{Ablation Study}

To verify the effectiveness of each module in our trader agent, we remove the profile and observation module in the agent as ablation models.
From the results shown in Table~\ref{tab:ablation}, we can find that when we remove the profile module and use the same simple trading instruction for each agent, we obtain the lowest turnover rate and volatility.
When all agents share the same trading strategy, their actions tend to converge. 
For example, if all agents decide to buy a particular stock at the same time, it becomes challenging to complete transactions, resulting in low turnover rates and volatility.

When we remove the observation module, the agent is unable to comprehensively understand the current market conditions, resulting in more random trading behaviors. 
This led to increased volatility. 
The lack of informed trading disrupted the strategic interactions typically present in the stock market, resulting in a higher order execution rate.

\begin{figure}[htbp]
    \centering
    \subfloat[The top shows the average stock return, and the bottom shows the order number.]{\includegraphics[width=0.51\textwidth]{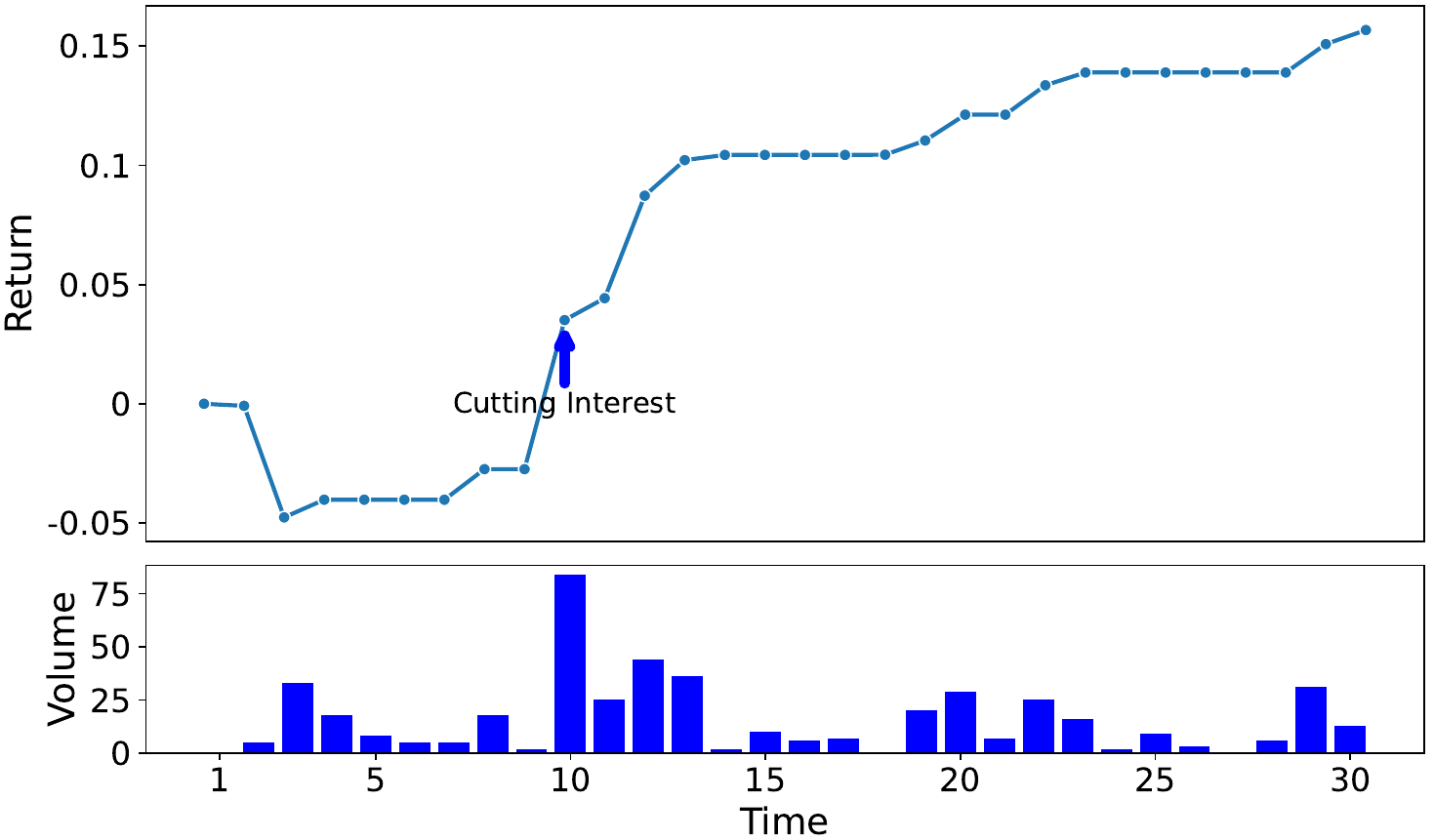}} 
    \subfloat[Stock price.]{\includegraphics[width=0.4965\textwidth]{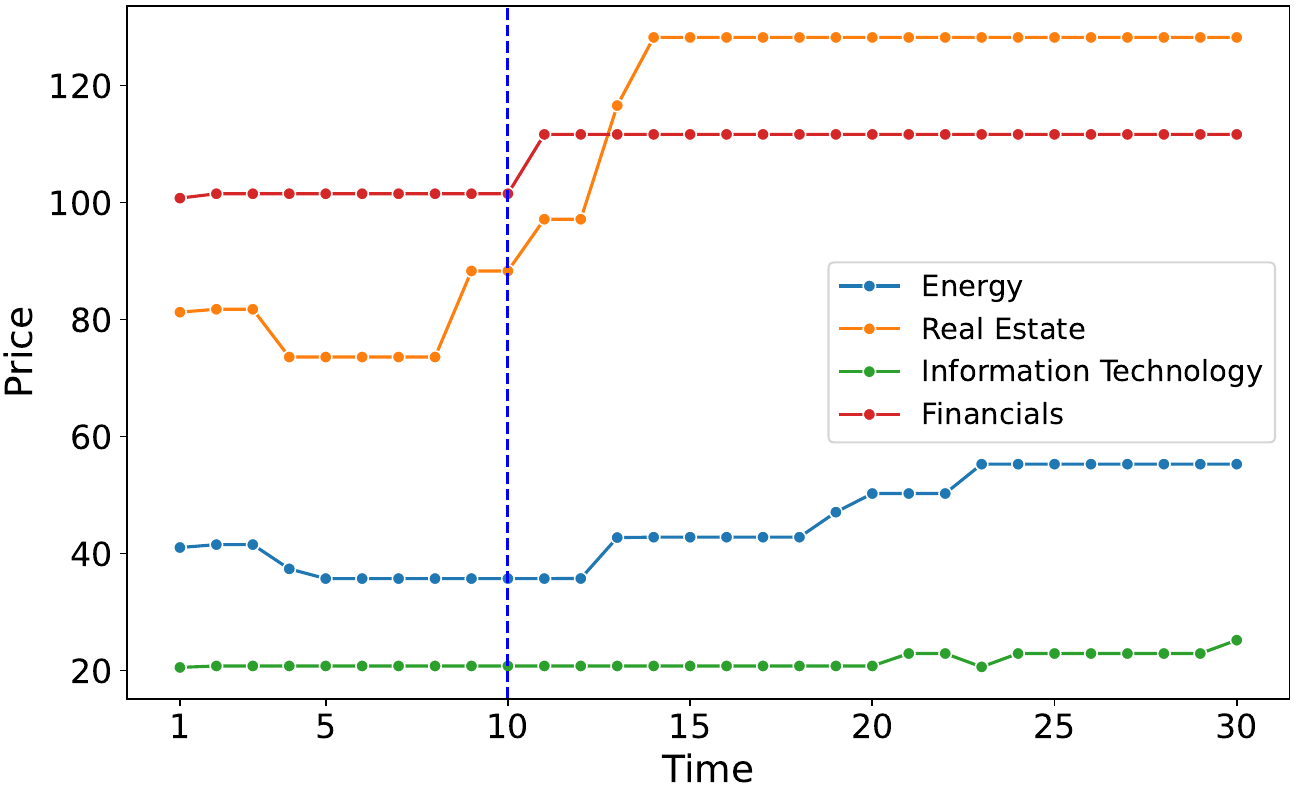}} 
    \caption{The impact of cutting interest rates policy shock.} 
    \label{fig:cut-rate}
\end{figure}

\begin{minipage}{\linewidth}
  \centering
  \begin{minipage}{0.47\linewidth}
      \begin{figure}[H]
          \includegraphics[width=\linewidth]{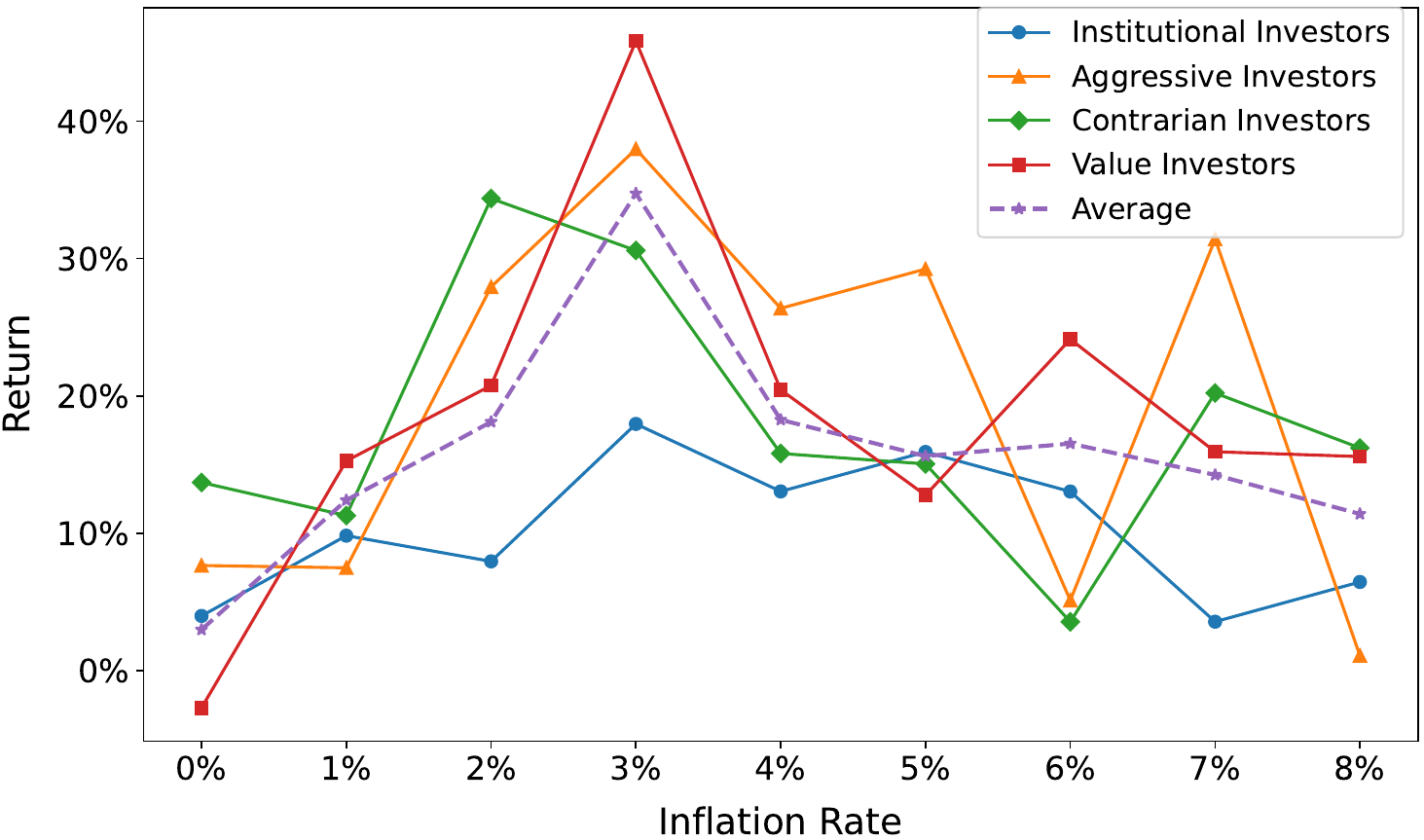}
          \caption{Stock return under different inflation rate.}
          \label{fig:inflation-agent}
      \end{figure}
  \end{minipage}
  \hspace{0.01\linewidth}
  \begin{minipage}{0.47\linewidth}
      \begin{figure}[H]
          \includegraphics[width=\linewidth]{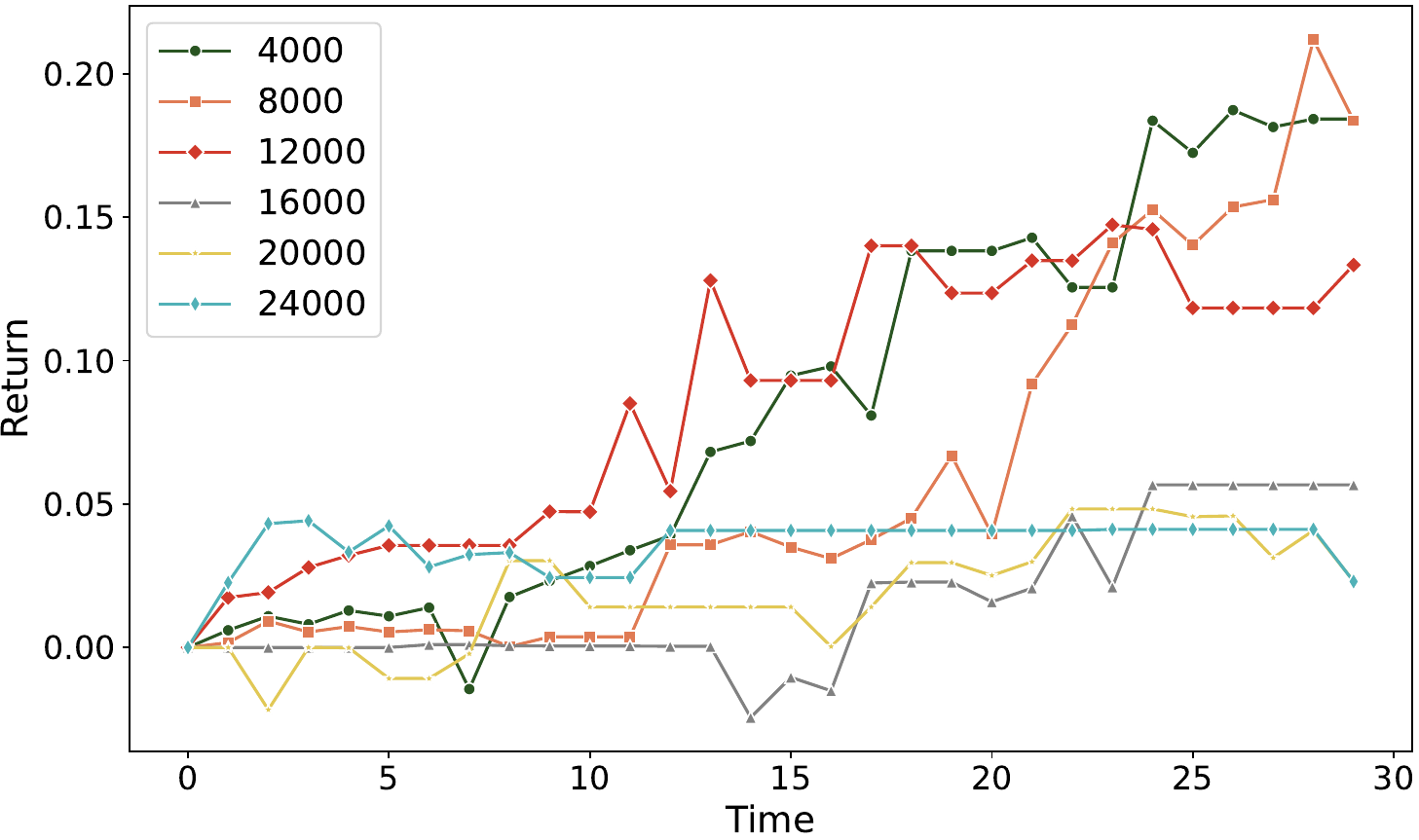}
          \caption{Average return for the agent with different initial capital.}
          \label{fig:large-trader}
      \end{figure}
  \end{minipage}
\end{minipage}

\subsection{Policy Shock Analysis: Influence of Cutting Interest Rates}

To verify the impact of a rate cut on the stock market, on the 10th day of normal trading, we informed each agent about the Federal rate cut by releasing news, such as ``According to the latest minutes of the Federal Open Market Committee (FOMC) monetary policy meeting, the Federal Reserve has decided to cut interest rates by 50 basis points, continuing to maintain the target range for the federal funds rate''.
Figure~\ref{fig:cut-rate}(a) illustrates the changes in returns for various agents following the announcement of the Federal rate cut.
As shown in Figure~\ref{fig:cut-rate}(a), after informing the simulated market about the rate cut, the average return of all stocks gradually increased.

This phenomenon observed in our simulation system aligns with current economic theories. According to contemporary economic viewpoints, a rate cut typically has a positive impact on the stock market because it lowers borrowing costs for businesses and consumers, stimulates consumption and investment, enhances economic growth expectations, and reduces the attractiveness of fixed-income products.


Furthermore, different industries respond differently to rate cuts. 
For instance, highly leveraged sectors like real estate and energy may benefit significantly due to their sensitivity to interest rate changes, as lower financing costs can substantially boost their profitability. 
Conversely, the financials (\eg banks) might experience a contraction in net interest margins, which could have some negative effects.
Figure~\ref{fig:cut-rate}(b) shows the stock price changes in the real estate and financials in our simulated market, which also reflect observations consistent with traditional economic theories.

\subsection{Influence of Inflation Shock}



In the experiment, we informed each agent of the current inflation rate on the first day by releasing news, such as ``In recent months, the national inflation rate has risen to a high point in the past decade, attracting widespread attention from the market and residents. According to data from the National Bureau of Statistics, the inflation rate has now reached 8.5\%''. 
After 30 trading days, we calculated the rate of return for each agent and the average return for all stocks in the simulated stock market. 
Figure~\ref{fig:inflation-agent} shows the return rate of various agents under different inflation shocks.
As shown in Figure~\ref{fig:inflation-agent}, both excessively high and excessively low inflation rates negatively impact the rate of return. 
Conversely, when the inflation rate is between 2-3\%, all the agents' returns and the average return of the overall stock market reach relatively high levels.

This phenomenon observed in our simulation system aligns with current economic theories. 
Contemporary economic research~\cite{gokal2004relationship, boudoukh1993stock} suggests that high inflation often indicates increased production and operational costs, which weaken consumer purchasing power and reduce demand. 
Additionally, high inflation is typically associated with increased economic uncertainty, leading to a decline in investor confidence.
Similarly, low inflation or deflation reflects weak demand in the economy. 
Low inflation is usually correlated with slower economic growth and reduced willingness of businesses to expand.
Generally, moderate inflation (around 2-3\%) is considered a hallmark of a healthy economy and typically has a positive impact on the stock market.

\subsection{LLM and Behavioral Economics: Exploring the Trader with Behavior Bias}

As shown in \S~\ref{sec:scenarios}, we altered the investor type ratios to explore the influence of behavior bias.  
We explore the market dynamics under two scenarios: (1) all participants in the market are value investors and (2) all participants are aggressive investors. 
Table~\ref{tab:ablation} presents the results for these two scenarios. 
As shown in Table~\ref{tab:ablation}, when the market consists entirely of value investors, the overall market volatility is lower by 0.1\% compared to the setting with a real market distribution. 
In contrast, when the market consists entirely of aggressive investors, there is a significantly higher order quantity and turnover rate, and the market volatility also shows a notable increase.

The modeling of behavioral economics has long perplexed economists~\cite{de1990noise,hirshleifer2020presidential} since the behavior patterns of irrational individuals are difficult to describe with rigorous mathematics. 
Consequently, when economic models include irrational individuals, these models are often difficult to formulate and solve. 
In contrast to rule-based microeconomic models, LLMs can better understand complex human behavior patterns than mathematical models.
Within the framework of LLM-based simulation framework presented in this paper, human behavioral characteristics and biases can be easily incorporated into economic models.

\subsection{Exploring the Large Trader Impact}

To explore the large trader impact, we initialize agents with different amounts of starting capital and observed the changes in their returns. 
As shown in Figure~\ref{fig:large-trader}, small traders generally exhibit a higher rate of return growth after 30 trading days, whereas large traders tend to show poorer return performance. 
Preliminary economic research~\cite{gastineau1991large,bank2004hedging,frey1998perfect} suggests that this phenomenon can be attributed to the large trader's substantial capital, which makes it difficult to find sufficient trading counterparts for their buy and sell orders. 
Consequently, the action of large trader is less flexible compared to small trader. 
From the results of this experiment, it demonstrate that \model maintains consistency with traditional economic conclusions even in scenarios with these indeterminate conclusions.
\section{Future Directions}

Although the \model can effectively simulate the operating principles of financial markets, it remains a pilot study using LLM-based agents in economic modeling. 
There are still many directions that need further exploration:

(1) More financial market participants and behaviors need to be incorporated, such as banks, loans, bond issuance, and a wider variety of asset types. 
In real-world financial markets, these elements are crucial components, and their impact on the economy should be considered.

(2) The operational status of listed companies needs to change with market fluctuations, which in turn affects their stock prices.

(3) More agent behaviors should be considered, such as employment, purchasing mutual funds, and buying real estate. 
These common factors may also influence personal assets.

\section{Related Work}


\paragraph{Language Model based Agent.}
Large language models have demonstrated rich world knowledge~\cite{zhang2023building,tang2023medagents}, complex logical reasoning abilities~\cite{chen2023reconcile}, and in-context learning capabilities~\cite{fu2023improving} in various natural language understanding and generation tasks. 
Recently, many studies have explored the construction of LLM-based agents by setting profiles, inputting knowledge as memory, and endowing them with the ability to call external tools~\cite{qin2023toolllm,qin2023tool}. 
Researchers have also proven the effectiveness of LLM-based agents in many tasks such as tool learning~\cite{zhuang2024toolqa,hsieh2023tool}, recommendation systems~\cite{huang2023recommender,zhang2024prospect}, and multi-modal tasks~\cite{yang2023appagent,yang2023mm}. 
Furthermore, some works have started to explore how to construct communication mechanisms among multiple agents, allowing them to collaborate on more complex tasks such as travel planning~\cite{xie2024travelplanner}, software development~\cite{qian2023communicative}, and scriptwriting~\cite{Zheng2023AgentsMO}.

\paragraph{Large Language Models in Economics Research.}
Large language models have been widely used in economics research. 
Researchers collect data of the financial domain to instruct models for many tasks such as sentiment analysis~\cite{lee2023stockemotions,shah2022flue}, text classification~\cite{li2023chatgpt}, named entity recognition~\cite{loukas2022finer}, and stock movement prediction~\cite{soun2022accurate}. 
In the task of stock movement prediction, \citet{lopez2023can} use ChatGPT to analyze sentiment trends and establish their correlation with stock market movements; 
\citet{xie2023pixiu} combines financial data and news to fine-tune large models, integrating numerical and textual features to predict stock prices; 
\citet{tong2024ploutos} assigns large models to play different experts, generating predictions from sentiment analysis, technical analysis, and history analysis, and synthesizing them to produce state-of-the-art stock price predictions. 
These works demonstrate that LLMs have a good understanding of financial knowledge, and are capable of tackling financial tasks.



Economic behavior is a complex form of social behavior, such as investing, employment, and taxation.
\cite{li2023large} proposed using LLM-based agents to simulate macroeconomic markets. 
They first constructed mathematical models for banks and governments, which generated various economic indicators. 
Then they provide these indicators to the agents, and the agents output consumption and employment indices and feedback into the bank and government models, thereby achieving a simulation of the macroeconomic market. 
However, under the traditional mathematical models paradigm, this method only used the LLM-based agent as a numerical indices prediction module.
In contrast, our \model represents a fully agent-based microeconomic model where all decisions are made by the agents based on their observations. 
Unlike traditional mathematical models, this approach offers a novel paradigm for modeling complex human behaviors in economic research.
\section{Conclusion}

In this paper, we present a pioneering approach \fullmodel (\model) to model the financial market by using LLM-based agents. 
To build a trader agent, we employ three modules to establish the stock market's trading ability, including the profile module with trading strategy instructions, the observation module to understand the dynamics of market and economic policies, and the tool-learning based action module.
By creating a simulated stock market with an actual trading mechanism, numerous agents with different trading strategies and initial capital can conduct trading in the \model.
To verify the consistency between the simulated market and the real stock market, we construct four scenarios including a federal interest rate cut, inflation shock, the impact of trader behavior bias, and large trader impact.
From these experiments, we observe that our \model demonstrates a sophisticated understanding of market dynamics and financial policy information. 
The agents make decisions based on their trading strategy, mirroring the complexity and variability of human behavior more accurately than traditional mathematical economic models.
And the \model offers a new perspective for a more dynamic and realistic simulation of financial markets research.



\clearpage
\bibliographystyle{unsrtnat}
\bibliography{reference}


\appendix

\section{Appendix}

\subsection{Instruction of Agent Trading Strategy}~\label{sec:appendix.A1}

The detailed trading preference description of each agent as follows:
\begin{itemize}
    \item \textbf{Value investor}: You are a value investor focused on identifying stocks whose market prices are below their intrinsic values. Your investment decisions are based on thorough financial analysis, seeking long-term stable returns rather than short-term price fluctuations. By examining the fundamentals of companies, such as profitability, financial health, and industry position, you can identify and invest in high-quality companies undervalued by the market. As a value investor, you patiently wait for the market to reassess the true value of these stocks, realizing the appreciation of your investments.
    \item \textbf{Institutional investor}: You are an institutional investor, typically representing a large investment company. You possess substantial capital and expertise, engaging in the purchase and sale of stocks, bonds, and other financial assets. Your investment decisions are based on in-depth market analysis, long-term financial planning, and sophisticated risk management strategies. Your primary goal is to secure stable and reliable returns for the institution or its beneficiaries you represent.
    \item \textbf{Contrarian investor}: You are a contrarian investor, specializing in finding stocks that are generally overlooked or undervalued by the market. You believe in buying when market sentiment is low and selling when it's high, aiming to profit from this approach.  Your strategy requires a high level of patience and steadfast determination, as well as a deep understanding of market psychology and fundamental analysis, to be greedy when others are fearful and thus achieve capital appreciation.
    \item \textbf{Aggressive investor}: You are an Aggressive investor, focusing on profiting from short-term market fluctuations. You prefer quick trades, using technical analysis and market trends to predict immediate price movements. Your investment strategy often involves higher risks, but it can also yield quicker returns. You need to constantly stay sensitive to market dynamics and be ready to enter and exit the market at any time to respond to rapidly changing market conditions.
\end{itemize}

\subsection{Detailed Definition of Companies}~\label{sec:appendix.company}

We show the detailed definition of the simulated listed company below:

\begin{itemize}
    \item \textbf{EN001}: An energy company primarily engaged in oil and gas extraction. Initial Prices: [10.00, 10.20, 10.50, 10.35, 10.60]
    \item \textbf{MA002}: A specialized chemical materials producer, whose products are widely used in construction and manufacturing. Initial Prices: [20.00, 19.85, 20.15, 20.50, 20.75]
    \item \textbf{IN003}: An industrial company providing mechanical equipment and automation solutions. Initial Prices: [30.00, 29.50, 30.25, 30.75, 31.00]
    \item \textbf{CC004}: A company specializing in high-end electronic consumer products, such as smartphones and laptops. Initial Prices: [40.00, 39.50, 40.25, 41.00, 41.50]
    \item \textbf{DC005}: A producer and seller of daily consumer goods, such as food and beverages. Initial Prices: [50.00, 49.75, 50.50, 50.25, 51.00]
    \item \textbf{HC006}: A healthcare company providing innovative medical devices and pharmaceuticals. Initial Prices: [60.00, 59.50, 60.25, 60.75, 61.50...]
    \item \textbf{FI007}: A provider of comprehensive financial services, including banking, insurance, and asset management. Initial Prices: [70.00, 70.50, 71.00, 71.50, 72.00]
    \item \textbf{IT008}: A leading software development and information technology services provider. Initial Prices: [80.00, 79.75, 80.50, 81.25, 81.75]
    \item \textbf{TS009}: An operator of extensive telecommunications networks, providing data and communication services. Initial Prices: [90.00, 90.50, 91.00, 91.50, 92.00]
    \item \textbf{UT010}: A utility company providing water, electricity, and natural gas services. Initial Prices: [100.00, 99.50, 100.25, 100.75, 101.50]
    \item \textbf{RE011}: A company primarily engaged in real estate development and management, covering commercial and residential projects. Initial Prices: [110.00, 110.50, 111.00, 111.50, 112.00]
\end{itemize}

\end{document}